\documentclass[twoside,11pt]{article}

\usepackage{jmlr2e}
\usepackage[noabbrev,capitalize]{cleveref}
\usepackage{textcomp}

\newcommand{\tvx}{\texttt{t5x}}
\newcommand{\seqio}{\texttt{seqio}}

\firstpageno{1}

\begin{document}

\title{Scaling Up Models and Data with \texttt{t5x} and \texttt{seqio}}

\author{\\
\email Lead Authors \\
\name Adam Roberts \email adarob@google.com \\
\name Hyung Won Chung \email hwchung@google.com \\
\name Anselm Levskaya \email levskaya@google.com \\
\name Gaurav Mishra \email mishragaurav@google.com \\
\name James Bradbury \email jekbradbury@google.com \\\\
\email Technical Contributors \\
\name Daniel~Andor, Sharan~Narang, Brian~Lester, Colin~Gaffney,
Afroz~Mohiuddin,\\ Curtis~Hawthorne,  Aitor~Lewkowycz, Alex~Salcianu,  Marc~van~Zee, Jacob~Austin, Sebastian~Goodman, Livio~Baldini~Soares, Haitang~Hu, Sasha~Tsvyashchenko, Aakanksha~Chowdhery, Jasmijn~Bastings, Jannis~Bulian, Xavier~Garcia \\ Jianmo~Ni, Andrew~Chen, Kathleen~Kenealy, Jonathan H. Clark, Stephan~Lee \\ Dan Garrette, James Lee-Thorp  \\\\
\email Technical Advisors \\
\name  Colin~Raffel, Noam~Shazeer, Marvin~Ritter,\ Maarten~Bosma, Alexandre~Passos, Jeremy~Maitin-Shepard \\\\
\email Leadership \\
\name Noah Fiedel, Mark Omernick, Brennan~Saeta, Ryan~Sepassi, \\Alexander~Spiridonov, Joshua~Newlan, Andrea Gesmundo
\\\\
\email *Authors are ordered by impact within groups.
}
\maketitle

\begin{abstract}
Recent neural network-based language models have benefited greatly from scaling up the size of training datasets and the number of parameters in the models themselves.
Scaling can be complicated due to various factors including the need to distribute computation on supercomputer clusters (e.g., TPUs), prevent bottlenecks when infeeding data, and ensure reproducible results.
In this work, we present two software libraries that ease these issues:
\tvx{} simplifies the process of building and training large language models at scale while maintaining ease of use, and \seqio{} provides a task-based API for simple creation of fast and reproducible training data and evaluation pipelines.
These open-source libraries have been used to train models with hundreds of billions of parameters on datasets with multiple terabytes of training data.
Along with the libraries, we release configurations and instructions for T5-like encoder-decoder models as well as GPT-like decoder-only architectures.

\tvx{} and \seqio{} are open source and available at \url{https://github.com/google-research/t5x} and \url{https://github.com/google/seqio}, respectively.
\end{abstract}


\section{Introduction}
Neural network models are highly scalable. In particular, Transformers \citep{vaswani2017attention} have been scaled up to hundreds of billions of parameters, demonstrating significant improvements on tasks of interest along the way. However, training models at these sizes is challenging and often demands specialized and hand-tuned software systems, making it difficult to quickly iterate over experimental research ideas. Furthermore, downstream usage and evaluation of these models requires either finetuning or prompting, which must be applied consistently across competing models.

These requirements motivate the development of a framework that enables easy model scaling while remaining research-friendly.
JAX~\citep{jax2018github,jax_paper} is uniquely positioned to provide such benefits; its NumPy-like~\citep{harris2020array} API makes it easy to understand and develop, while the \texttt{jax.pjit} API backed by XLA GSPMD \citep{xu2021_gspmd} provides a powerful and efficient compiler-based programming model for parallelism. In this paper, we present \tvx{}, a JAX-based open-source library that is focused on building Transformer models at a wide range of scales.

As model sizes grow, it becomes increasingly important to train them on larger datasets.
We therefore additionally introduce \seqio{}, a fully-featured library for efficiently managing data pipelines and model evaluation with a simple, task-based API. \seqio{} builds off of \texttt{tensorflow.data} with additional support for SPMD-based data parallelism, making it compatible with many modeling frameworks including JAX, TensorFlow~\citep{tensorflow2015-whitepaper}, and PyTorch~\citep{paszke9015_pytorch}. \seqio{} includes an option to create deterministic data pipelines, which we have found to be an indispensable tool for optimizing training performance, fairly comparing models, and diagnosing issues during training.

\section{\tvx{}}
\label{sec:t5x}

\tvx{} is a library for training, evaluating, and inferring with JAX models across many scales, with a focus on Transformer-based language models. The typical usage involves either pretraining from scratch or finetuning an existing language model implemented in Flax---a JAX-based neural network library \citep{flax2020github}---and then running inference for evaluations and/or downstream applications. GPU and CPU acceleration are supported, but \tvx{} is optimized for TPU.

In the following subsections, we discuss the design of \tvx{} including how it wraps \texttt{jax.pjit} to provide a high-level interface to XLA GSPMD for simple yet efficient scaling via parameter, activation, and data partitioning.

\subsection{Overall structure}
\vspace{-1ex}
\begin{figure}[htb]
  \centering
    \includegraphics[width=0.95\textwidth]{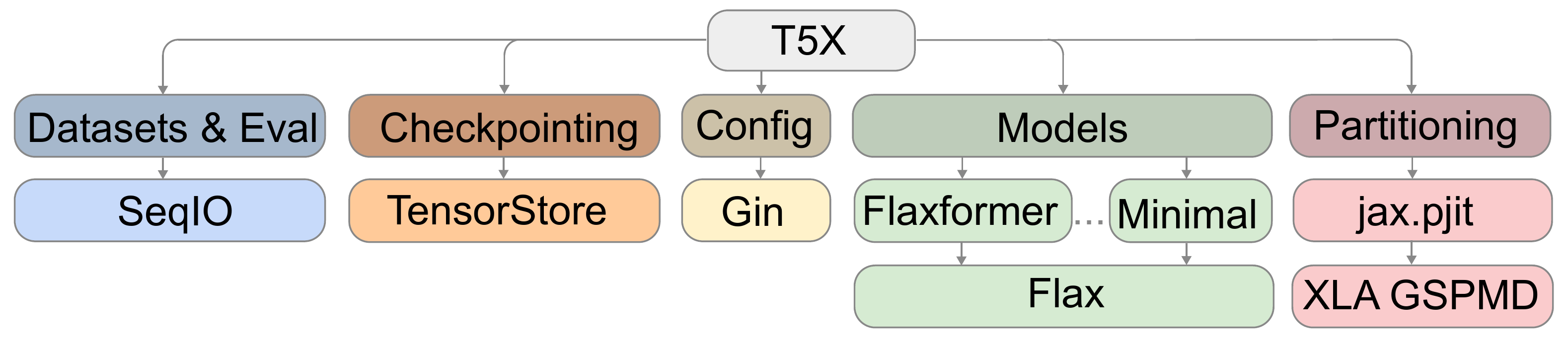}
  \caption{Overall structure of \tvx{}, showing which dependencies are used to implement the principal functionalities contained in darker boxes.}
  \vspace{-1ex}
  \label{fig:t5x_diagram}
\end{figure}

\cref{fig:t5x_diagram} illustrates the modular structure of \tvx{}, in particular how \tvx{} uses open-source libraries to implement different functionalities. We briefly describe each of these below.
\newpage
\begin{itemize}
    \item \textbf{Datasets and Evaluation} - By default, we use \seqio{} to create reproducible ``tasks”, which we cover in detail in \cref{sec:seqio}. Note that \tvx{} provides a modular structure and may be used without \seqio{}.
    \item \textbf{Checkpointing} - For large models, straightforward tasks like checkpointing can be challenging, especially when using parameter and optimizer partitioning. In order to efficiently manage checkpoints from multiple hosts with distributed parameters, we built our own checkpointing library utilizing \texttt{TensorStore}\footnote{\url{https://github.com/google/tensorstore}} as a tool for scalably reading and writing sliced tensors.
    \item\textbf{Configuration} - For fast iterations over research ideas, it is important for the codebase to be easily configurable. In particular, researchers should be able to control function arguments and even use custom components without needing to modify the core library code. We use Gin\footnote{\url{https://github.com/google/gin-config}} for this dependency injection. As a typical example, users can inject hyperparameters or a custom model object as function arguments for training. More advanced users can replace entire modules (e.g., a custom checkpointer) in a similar manner.
    \item\textbf{Models} - To actually implement the modeling layers, we use Flax~\citep{flax2020github}, a high-level library built on JAX. \cref{sec:model} discusses a few specialized features in Flax that are required by \tvx{} to enable model parallelism. 
    \item\textbf{Partitioning} - One of the key features of \tvx{} is its ability to parallelize over data, parameters, and activations. We use the XLA GSPMD partitioner~\citep{xu2021_gspmd} to automatically shard the computation graph and use \texttt{jax.pjit} as a frontend to interact with GSPMD, providing our own high-level API to simplify configuration. \cref{sec:partitioning} will discuss this component in further detail.
\end{itemize}

\newpage
\subsection{XLA GSPMD partitioning with \texttt{jax.pjit}}
\label{sec:partitioning}

Many different kinds of parallelism are useful for scaling large models. In data parallelism, the input data and intermediate activations are split over devices (``partitioned'') along the global batch axis. In (tensor) model parallelism, the model computation for a single example, and the model parameters themselves, are split across devices.

These two kinds of parallelism are orthogonal, in that a system with $N=M \times D$ devices can use $M$-way model parallelism and $D$-way data parallelism at the same time. \tvx{} assumes such a decomposition of the devices in the system into a model parallel axis (specified in multidimensional TPU networks as a ``model parallel submesh'') and a data parallel axis or submesh.

\tvx{} has built-in support for multiple variants of data and model parallelism, i.e., multiple ways to use these two orthogonal axes:
\begin{itemize}
    \item Data parallelism is parallelism in the batch dimension. It can also involve either replicating the parameters and optimizer state or sharding them over the data parallel axis. The former is also termed ``1D parameter partitioning'', since parameters are only subject to model parallel partitioning over one array axis, while the latter is ``2D parameter partitioning'' since a second array axis in each parameter is also partitioned.
    \item Model parallelism involves partitioning model computation over axes other than the batch dimension. In Transformers, it involves partitioning parameters and some intermediate activations along axes like the MLP hidden dimension and the heads dimension. Other intermediate activations (those with an embedding/model axis but not hidden/heads) can either be replicated over the model parallel axis (``1D activation partitioning'') or sharded (``2D activation partitioning'').
\end{itemize}
These options correspond to previously described parallelism techniques: 2D parameter partitioning is also known as ZeRO-3 \citep{rajbhandari2019zero} or fully sharded data parallelism; 1D activation partitioning is also known as Megatron \citep{shoeybi2019megatron} and is the default in the Mesh TensorFlow Transformer \citep{shazeer2018_mesh_tensorflow}; and 2D activation partitioning is the ``fully sharded'' case described in \cite{xu2021_gspmd}.

\tvx{} supports flexible partitioning configurations, including these built-in options, using the Flax APIs described in the following section.

\subsection{Model Implementation}
\label{sec:model}

\tvx{} is compatible with Flax-based model implementations with some minor caveats. In order to support flexible configurations for parameter and activation partitioning, \tvx{} requires these tensors to be annotated with user-defined logical named axes when they are defined in the model implementation, via \texttt{flax.partitioning.param\_with\_axes}. These logical axes are used to group tensor dimensions that one would expect to always partition in the same way in various settings, for example ``batch'' (for partitioning across examples in a batch), ``kv'' (for partitioning across the dimensions of key-value matrices in Transformer self-attention layers), or ``head'' (for partitioning across heads in multi-headed attention). While XLA GSPMD will automatically select matching partitions for the intermediate activations produced by these parameters, users may also provide overrides by naming their axes as well via \texttt{flax.partitioning.with\_sharding\_constraint} to better optimize memory usage and between-device communication.

At runtime, the user provides a mapping from each logical axis name to one of the two hardware axes (\texttt{model} and \texttt{data}). Alternatively, the logical axis can be mapped to \texttt{None} to specify that it should be replicated across all devices.

Given a \texttt{flax.nn.module} module implemented as described above, one must simply wrap it in a subclass of  \texttt{t5x.BaseModel} to define its loss, evaluation, and inference methods to make it compatible with the core \tvx{} interface.

With its modular design, the model implementations in \tvx{} can be flexible. Layers and modules can be written directly with Flax (e.g., the ``Minimal'' implementations discussed in \cref{sec:example_models}) or using a higher-level library such as Flaxformer\footnote{\url{https://github.com/google/flaxformer}}. Dependency injection with Gin allows users to easily swap the module implementation in their configuration. Even when implemented in different libraries, the model checkpoints can be made compatible. Additionally, models trained with the legacy T5 codebase\footnote{\url{https://github.com/google-research/text-to-text-transfer-transformer}} based on Mesh TensorFlow can be read directly by \tvx{}. They can also be converted to the native \tvx{} format resulting in faster reading based on how \tvx{} leverages TensorStore.

\section{\seqio{}}
\label{sec:seqio}

\seqio{} is a library for processing data to be fed into models for training, inference, and evaluation. It uses \texttt{tensorflow.data} to create scalable data pipelines but requires minimal use of TensorFlow. In particular, with one line of code, the returned dataset can be transformed to a NumPy iterator and hence it is fully compatible with other frameworks such as JAX or PyTorch.

\subsection{Task-based API}

\begin{figure}[h]
  \centering
    \includegraphics[width=0.5\textwidth]{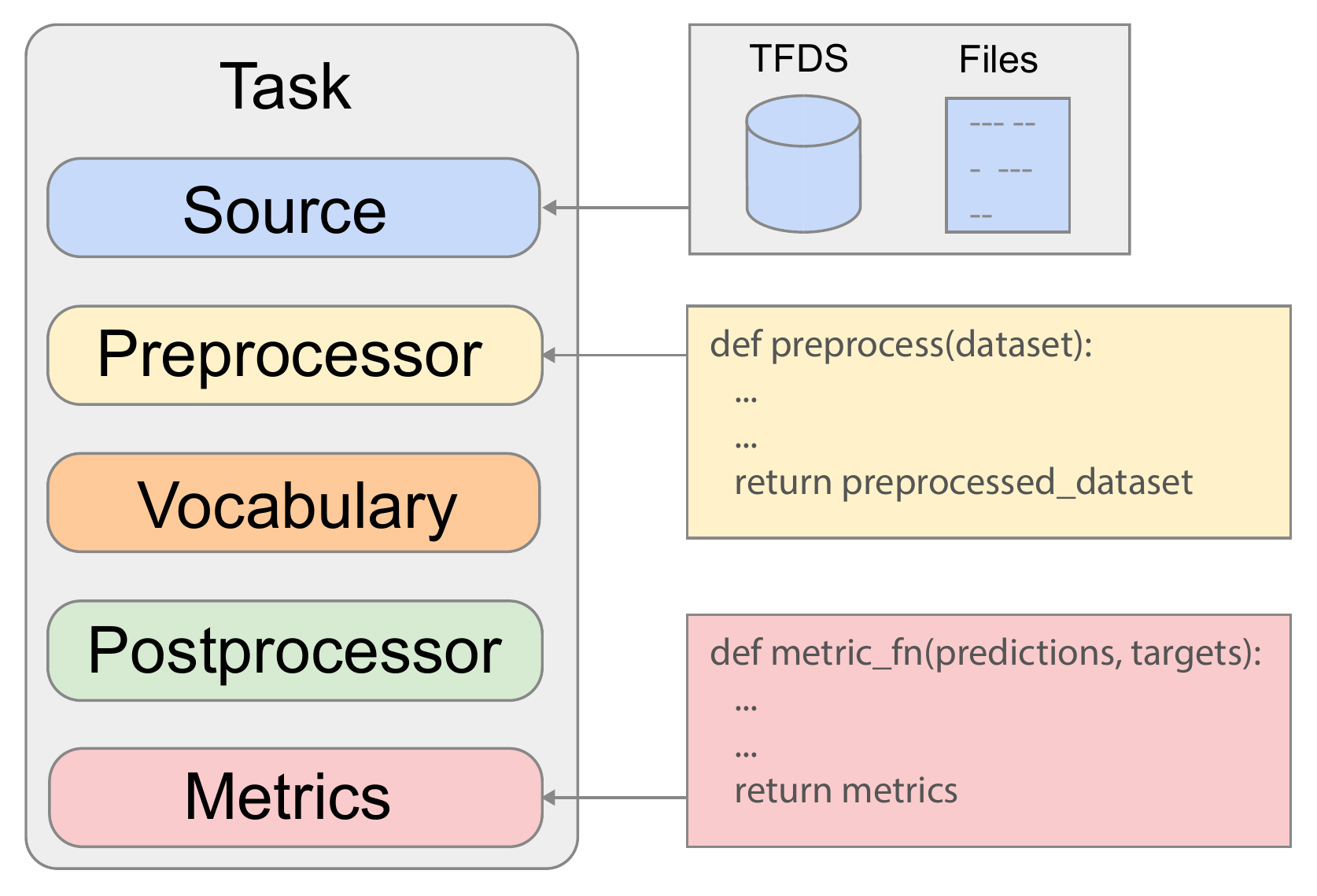}
  \caption{Structure of a \seqio{} Task, highlighting customizable use of APIs.}
  \vspace{-1ex}
  \label{fig:seqio_diagram}
\end{figure}

A key differentiator of \seqio{} from most other dataset frameworks is its use of a task-based API, which is illustrated in \cref{fig:seqio_diagram}. The Task object associates raw data sources with preprocessing steps---to define the inputs and targets---and evaluation metrics---to create consistent benchmarks. Feature converters are used to convert task features into the raw values that will be fed into the model itself. This way the same task can be made compatible with various architectures (e.g., encoder-decoder or decoder-only).

Multiple Tasks can also be combined into a Mixture for multi-task training, with user-provided mixing rates.

\subsection{Deterministic Pipelines}
\label{sec:deterministic_pipelines}
\seqio{} also provides the ability to generate deterministic data pipelines with following key features: 

\begin{itemize}
    \item\textbf{Reproducibility} - Data read from a deterministic Task/Mixture is always in the same order, which is important for dataset debugging and inspection. It also allows the dataset to be held constant for more accurate benchmarking or when debugging training issues.
    \item\textbf{Recoverability} - A deterministic dataset can be continued from an arbitrary point in training. This avoids repeating data on intentional or unintentional (e.g., due to preemption) training restarts, which can lead to reduced performance and memorization \citep{lee2022dedup}. We have also found it useful for manually skipping batches of a dataset that produce instabilities during training.
    \item\textbf{Sharding} - Data can be arbitrarily sharded across any number of readers to enable efficient distributed reads from data-parallel workers.
    \item\textbf{Global Shuffling} - Deterministic datasets are prepared by an offline job that ensures data is well-shuffled. This is particularly important when examples are correlated (e.g., they are based on the same source document) or multiple epochs are used.
\end{itemize}

To achieve these requirements, once a Deterministic Task/Mixture is defined, a distributed caching job (implemented in Apache Beam\footnote{\url{https://beam.apache.org}}) loads the raw data, preprocesses and shuffles the examples, assigns ordered indices, and writes the data to sharded files. Importantly, the examples are sharded by the modulo of their index to the number of files. This enables each set of data parallel hosts to sequentially read an interleave an exclusive set of files at train time, helping to optimize throughput and greatly reducing the chance of an input bottleneck.

We have found these four features to be beneficial when training extremely large models. They increase throughput, protect against overfitting, ease debugging, and provide fine-grained control over the examples seen during training in order to avoid instabilities. We expect this control to also be useful for researchers interested in understanding how specific aspects of the dataset (e.g., order and repeats) might affect the model's ability to generalize or memorize.





\newpage

\section{Example Models}
\label{sec:example_models}

With \tvx{}, we provide well-tested\footnote{We validated these models by reproducing the T5 models from~\citet{t5_paper}, originally implemented in Mesh TensorFlow Transformer.} ``Minimal'' model implementations with checkpoints:
\begin{itemize}
    \itemsep0em 
    \item T5 from~\citet{t5_paper} and T5.1.1 (introduced after the paper).
    \item Scalable T5
    \item mT5~\citep{xue-etal-2021-mt5}
    \item ByT5~\citep{xue2021byt5}
\end{itemize}
These model implementations are minimal in the sense that they only use Flax with limited abstractions as opposed to using higher-level libraries built on top of Flax (e.g. Flaxformer). They closely mimic the pedagogical Flax examples\footnote{\url{https://github.com/google/flax/tree/main/examples}}.

Scalable T5 is an implementation of T5.1.1 using \texttt{jax.scan} to significantly reduce compilation time and provide finer-grained control over activation memory.

We also provide a model configuration (without any checkpoints) for a decoder-only architecture that is compatible with LaMDA~\citep{thoppilan2022_lamda}.


\section{Related Work}
There are many open source libraries for training sequence models. Previous Google-released systems based on TensorFlow include Tensor2Tensor~\citep{vaswani2018tensor2tensor}, Lingvo~\citep{shen2019lingvo}, and the Mesh TensorFlow~\citep{shazeer2018_mesh_tensorflow}-based T5 \citep{t5_paper}.

Comparable projects from other research groups include model libraries such as fairseq \citep{ott2019fairseq}, large-scale parallelism libraries such as FairScale~\citep{FairScale2021}, and libraries that include both kinds of functionality such as DeepSpeed~\citep{rasley2020_deepspeed} and Megatron~\citep{smith2022_megatron}.

Some major differentiators of \tvx{} are its use of JAX and Flax for model expression, its support for TPU (including TPU v4), and its Gin-based configuration system that allows uses to modify nearly everything about the model and training procedure. \tvx{} also doesn't support pipeline parallelism, a major component of systems like DeepSpeed. This is because the inter-chip network of TPUs has performance similar to within-node GPU interconnects but scales to thousands of chips, so model and data parallelism are sufficient to train efficiently at large scale.

\section{Project Status and Adoption}

We started the project in the fall of 2020 and open sourced the library code in October 2021. During that time, \tvx{} and \seqio{} achieved widespread adoption by teams across Google: \tvx{} has been launched on TPU hundreds of thousands of times at Google, and the total number of internal \tvx{} and \seqio{} users exceeds 1,000. Teams are using these libraries for research projects (from small-scale research to the largest language models trained at Google) and user-facing products. External adopters include academic and commercial users of Cloud TPUs, such as portions of the the Big Science project \citep{bigscience2022t5x}.

Users of \tvx{} and \seqio{} cite the usability and research-friendliness of the libraries as reasons for adoption. We are continuing to actively develop both libraries, prioritizing future work based on researcher needs and feedback.

\section{Contributions}

Adam Roberts founded and leads the project, designed and wrote much of \seqio{} and \tvx{}, and co-authored the paper. Hyung Won Chung designed and wrote much of \tvx{}, led its open sourcing, and co-authored the paper. Anselm Levskaya built the initial prototype for \tvx{} and wrote much of the code. Gaurav Mishra leads \seqio{}, implemented deterministic pipelines, and co-authored the paper. James Bradbury implemented partitioning in \tvx{} and co-wrote the paper.

Daniel Andor, Sharan Narang, Brian Lester, Colin Gaffney, Afroz Mohiuddin, Curtis Hawthorne,  Aitor Lewkowycz, Alex Salcianu, Marc van Zee, Jacob Austin, Sebastian Goodman, Livio Baldini Soares, Haitang Hu, Sasha Tsvyashchenko, Aakanksha Chowdhery, Jasmijn Bastings, Jannis Bulian, Xavier Garcia, Jianmo Ni, Andrew Chen, Kathleen Kenealy, Jonathan H. Clark, Stephan Lee, Dan Garrette, and James Lee-Thorp made substantial code contributions. 

Colin Raffel and Noam Shazeer helped design \seqio{}. Marvin Ritter advised on deterministic pipelines and the use of CLU Metrics. Maarten Bosma helped design deterministic pipelines. Jeremy Maitin-Shepard advised on the use of TensorStore. Alexandre Passos and Ryan Sepassi advised on overall technical design.

Noah Fiedel is a member of the leadership team, contributed to the high level design and roadmap, and co-wrote the paper. Mark Omernick, Brennan Saeta, Ryan Sepassi, Alexander Spiridonov (Product Manager), and Josh Newlan (Technical Program Manager) are members of the leadership team and co-wrote the paper. Andrea Gesmundo is a member of the leadership team and contributed to the internal infrastructure component.

Thanks to the many other contributors to the project: Ian Simon, Reiner Pope, Vincent Zhao, Pierre Ruyssen, Linting Xue, Junwhan Ahn, Barret Zoph, David Dohan, Masumi Parekh, Chang Lan, Frederick Liu, Julien Amelot, Luheng He, Fede Lebron, Rebecca Chen, Anosh Raj, Mandy Guo, Ethan Dyer, Mihai Tiuca, Hongkun Yu, Kevin Brooks, David Soergel, Kelvin Guu, Joshua Ainslie, Luyao Xu, Ji Ma, Josh Gardner, Daphne Ippolito, Peter Hawkins, Bo Pang, Marc Rasi, Wei Li, Wenhu Chen, Iulia Turc, John Wieting, Alex Passos, Zonglin Li, Katie Everett, Marvin Ritter, Olivier Bachem, Francesco Piccinno, Jakub Adamek, Jonathan Heek, Parker Schuh, Hexiang  Hu, Du Phan, Max Moroz, David Miller, Ryan Doherty, David Elworthy, Alfonso Castaño, Julian Eisenschlos, Vlad-Doru Ion, Lucas Dixon, Ron Shapiro, Dinghua Li, Aaron Parisi, Xi Chen, Nan Ding, Chung-ching Chang, Timothy Dozat, Natalia Ponomareva, Delesley Hutchins, Ankush Garg, Yu-Han Liu, Mehrdad Khatir, Costanza Conforti, Philipp Keck, Raphaël Marinier, Marie Pellat, Raghuram Vadapalli, Joshua Maynez, Yi Tay, Xihui Wu, David Belanger, Luke Metz, Dan Zheng, Deepti Bhatia, Hariharan Shanmugavadivel, Rewon Child, Rigel Swavely, Mihir Sanjay Kale, Arash Afkanpour, Roberto Rama, Juro Gottweis, Jonathan Herzig, Yilei Yang, Elias Mizan, Pedram Pejman, Jiayu Ye, Smit Sanghavi, Rahul Joshi, Ziqiang Feng, Charles Sutton, Weikang Zhou, Liam Fedus, Shanqing Cai, Ginger Perng, Yash Katariya, Urvashi Khandelwal, Sebastian Gehrmann, Edward Loper, Tianze Shi, Luke Vilnis, Amelia Archer, Tom Weingarten, David Zats, Murtaza Dhuliawala, Xin Xie, Sahil Dua, André Susano Pinto, Piotr Padlewski, Sascha Rothe, Erik Aas, Felix Stahlberg, Ken Durden, Christina Sorokin, Jaehoon Lee, Roy Frostig, Jacob Devlin, Jorge Gonzalez Mendez, Deepak Ramachandran, Santiago Ontanon, Karthik Raman, Yi Sun, Ali Elqursh, Reuben La Haye, Adam Fahrenkopf, Alex Polozov, Vinay Ramasesh, Ian Tenney

Thanks to Douglas Eck and Zoubin Ghahramani for sponsoring the project.


\vskip 0.2in
\bibliography{sample}

\end{document}